\documentclass[conference]{IEEEtran}
\IEEEoverridecommandlockouts
\usepackage{cite}
\usepackage{amsmath,amssymb,amsfonts}
\usepackage{graphicx}
\usepackage{textcomp}
\usepackage{xcolor}
\def\BibTeX{{\rm B\kern-.05em{\sc i\kern-.025em b}\kern-.08em
    T\kern-.1667em\lower.7ex\hbox{E}\kern-.125emX}}

\usepackage{algorithm}
\usepackage{algpseudocode}
\usepackage{tabularx}
\usepackage{booktabs}
\usepackage{multirow}
\newtheorem{theorem}{Theorem}

\usepackage[T1]{fontenc}
\usepackage{aecompl}
\usepackage[pagebackref=false,breaklinks=true,colorlinks=false,bookmarks=false,citecolor=blue,linkcolor=blue]{hyperref}
\usepackage[singlelinecheck=false]{caption} 

\newcommand*{\QEDB}{\hfill\ensuremath{\square}} 

\begin{document}
\title{Structured Pruning for Efficient ConvNets via Incremental Regularization \\
}

\vspace{-1em}

\author{
	\IEEEauthorblockN{
		Huan Wang$^1$, Qiming Zhang$^2$, Yuehai Wang$^1$, Lu Yu$^1$, Haoji Hu$^{1\dagger}$\thanks{$\dagger$ Corresponding author. This work is supported by the Natural Key R\&D Program of China (Grant No.~2017YFB1002400), Natural Science Foundation of Zhejiang Province (Grant No.~LY16F010004) and Chongqing Research Program of Basic science and Frontier Technology (cstc2016jcyjA0542).}
	}
	\IEEEauthorblockA{
		\textit{$^1$College of Information Science and Electronic Engineering},
		\textit{Zhejiang University},
		Hangzhou, China
	}
	\IEEEauthorblockA{
		\textit{$^2$School of Computer Science, FEIT},
		\textit{University of Sydney},
		Sydney, Australia\\
		\{huanw, wyuehai, yul, haoji\_hu\}@zju.edu.cn, qzha2506@uni.sydney.edu.au
	}
}
\vspace{-2em}


\maketitle
\begin{abstract}
%
Parameter pruning is a promising approach for CNN compression and acceleration by eliminating redundant model parameters with tolerable performance degrade.
Despite its effectiveness, existing regularization-based parameter pruning methods usually drive weights towards zero with \emph{large and constant} regularization factors, which neglects the fragility of the expressiveness of CNNs, and thus calls for a more gentle regularization scheme so that the networks can adapt during pruning. 
To achieve this, we propose a new and novel regularization-based pruning method, named \emph{IncReg}, to \emph{incrementally} assign different regularization factors to different weights based on their relative importance.
Empirical analysis on CIFAR-10 dataset verifies the merits of IncReg. 
Further extensive experiments with popular CNNs on CIFAR-10 and ImageNet datasets show that IncReg achieves comparable to even better results compared with state-of-the-arts. Our source codes and trained models are available here: {\color{black}{\href{https://github.com/mingsun-tse/caffe\_increg}{https://github.com/mingsun-tse/caffe\_increg}}}.
\end{abstract}

\section{Introduction}
Convolutional Neural Networks (CNNs) have made a remarkable success in computer vision tasks such as classification, detection, and segmentation by leveraging large-scale networks learning from a big amount of data. 
However, CNNs usually lead to massive computation and storage consumption, hindering their deployment on mobile and embedded devices. 
To reduce computation cost, many research works focus on the model compression and acceleration of CNNs.

Parameter pruning is a promising approach for CNN compression and acceleration, which aims at eliminating redundant model parameters with tolerable  performance degrade. 
One problem of parameter pruning is that it often produces unstructured and random connections which is hard to implement for speedup on general hardware platforms~\cite{HanLiuMao16}. 
To resolve this problem, many works focus on structured pruning which can shrink a network into a thinner one so that the implementation of the pruned network is efficient~\cite{AnwSun16,SzeCheYanEme17}.

There are mainly two categories of methods for structured pruning. 
One is importance-based methods, which prune weights in groups based on some established importance criteria~\cite{lecun1990optimal,HasSto93,MolTyrKar17}.  
The other is regularization-based methods, which add group regularization terms to the objective function and prune the weights by minimizing the objective function~\cite{VadLem16,wen2016learning,he2017channel} during training.

Existing group regularization approaches tend to use a \emph{large and constant} regularization factor for all weight groups in the network~\cite{wen2016learning,VadLem16}, which has two problems.  
First, this "one-size-fits-all" regularization scheme has a hidden assumption that all weights in different groups are equally important, which however does not hold true, because intuitively, weights with larger magnitude tend to be more important than those with smaller magnitude. 
Second, few works have noticed that the expressiveness of CNNs is so fragile~\cite{yosinski2014transferable} during pruning that it probably cannot withstand a large penalty term from the beginning.
Recently, AFP~\cite{DinDinHanTan18} is proposed to solve the first problem, ignoring the second one.
To resolve the second problem, in this paper, we propose a new regularization-based method named IncReg to \emph{incrementally} learn structured sparsity.
Our contribution in this work can be summarized into three folds:
\begin{itemize}
\item A new and novel progressive structured pruning method is proposed for CNN acceleration, which has empirically proven rather effective on popular deep neural networks compared with state-of-the-arts. 
\item The proposed pruning method has a relatively solid theoretical basis to support the intuition behind.
\item The proposed incremental regularization scheme brings more benefits, such as (1) robustness to hyper-parameter changes, (2) enabling the network to adapt during pruning, which empirically proves very valuable when pruning compact networks (\emph{e.g.}, ResNet~\cite{HeZhaRenSun16}) and pruning a large proportion of parameters.
\end{itemize}

\section{Related Work}
Parameter pruning enjoys a long history in the development of neural networks~\cite{Ree93}, which can mainly be categorized into two groups, \emph{i.e.}, importance-based and regularization-based.  
Importance-based pruning methods prune weights in groups based on some established importance criteria.  
For example, Optimal Brain Damage (OBD)~\cite{lecun1990optimal} and Optimal Brain Surgery (OBS)~\cite{HasSto93} propose an importance criteria based on the second-order derivatives of the loss function derived from Taylor expansion.  
Deep Compression~\cite{HanMaoDal15,HanTra15} prunes small-magnitude weights and obtains $9\sim 13\times$ parameter reduction on AlexNet and VGG-16.  
Taylor Pruning~\cite{MolTyrKar17} also derives a new importance criteria based on Taylor expansion, but they use the first-order derivatives considering their easy access during back-propagation.  
Their method is shown effective to prune filters on transfer learning tasks.  
\cite{LiKadDurEtAl17} uses $L_1$-norm to guide one-shot filter pruning, which proves effective on CIFAR-10 and ImageNet with VGG-16 and ResNet, but the reported speedup is very limited.  
Channel Pruning~\cite{he2017channel} alternatively uses LASSO-regression-based channel selection and feature map reconstruction to prune filters, and achieves one of the state-of-the-arts on VGG-16.  
AMC~\cite{he2018amc} is lately proposed to augment Channel Pruning with an optimized pruning ratio combination for different convolutional layers based on searching via reinforcement learning.  
Among regularization-based methods, Group-wise Brain Damage~\cite{VadLem16} and Structured Sparsity Learning~\cite{wen2016learning} embed Group LASSO~\cite{Yuan2006Model} into CNN regularization and obtain regular-shape sparsity.  
Both works assign a constant regularization factor to all the weight groups in the network.  
To the best of our knowledge, one recent work~\cite{DinDinHanTan18} proposes a regularization-based method by assigning different regularization factors to different weight groups, but contrary to our method, their assignment scheme is still constant, which does not take into account the dynamics of training process and the fragility of CNN expressiveness.
Besides, the effectiveness of their method is not tested on large-scale datasets.
Pruning is usually interpreted as the magnitude reduction of weights, however, inspired by dropout~\cite{HinSriKri12,srivastava2014dropout}, SPP~\cite{wang2017structured} recently is proposed to interpret pruning in a probabilistic manner with encouraging experimental results.
In a similar vein to SPP, several Bayesian pruning methods are proposed for both compression~\cite{louizos2017bayesian,molchanov2017variational} and acceleration~\cite{neklyudov2017structured}. 
Despite the impressive performance on relatively small datasets (\emph{e.g.}, MNIST, CIFAR-10), their effectiveness and flexibility on large-scale datasets (\emph{e.g.}, ImageNet) is still in question.

Apart from parameter pruning, there are primarily four other kinds of methods for CNN model compression and acceleration, including designing compact architectures, parameter quantization, matrix decomposition, and knowledge distillation. 
(1) Compact architecture design methods target more efficient and compact neural network architectures. 
For example, SqueezeNet~\cite{IanMosAsh16} is proposed to stack compact blocks, reducing the number of parameters of AlexNet by $50$ times. 
MobileNet~\cite{Howard2017MobileNets,sandler2018mobilenetv2} and ShuffleNet~\cite{Zhang2017ShuffleNet,ma2018shufflenet} leverage separable convolution operations to design networks for mobile applications. 
(2) Parameter quantization reduces CNN complexity by quantizing the weights and using fewer representation bits. 
\cite{CheWilTyrWeiChe15} proposes a hash function to group weights of each CNN layer into different hash buckets for parameter sharing. 
As the extreme form of quantization, binarized networks~\cite{CouBen16,LinCouMemBen16,RasOrdRedFar16} propose to learn binary weights or activations. 
%
%
(3) Matrix decomposition decomposes large matrices into several small matrices to reduce computation. 
\cite{DenZarBruLecFer14} shows that the weight matrix of a fully-connected layer can be compressed via truncated SVD. 
Several methods based on low-rank decomposition of convolutional kernel tensors are also proposed to accelerate convolutional layers~\cite{LebYarRakOseLem16,ZhaZouHeSun16}.
(4) Knowledge distillation transfers the learned knowledge from a large teacher model (or ensemble of models) to a small student model, which is pioneered by~\cite{buciluǎ2006model,ba2014deep} and refined by Hinton \emph{et al.}~\cite{hinton2015distilling}.
Ever since, various definitions of knowledge such as attention~\cite{zagoruyko2016paying} and metric structure~\cite{chen2018darkrank} have
been proposed to transfer the network expressiveness.

\section{The Proposed Method}
Consider a convolutional kernel, modeled by a 4-D tensor~$\mathbf{W}^{(l)} \in \mathbb{R}^{N^{(l)} \times C^{(l)} \times H^{(l)} \times W^{(l)}}$, where~$N^{(l)}$, $C^{(l)}$, $H^{(l)}$ and~$W^{(l)}$ are the dimension of the~$l$th ($1 \leq l \leq L$) weight tensor along the axis of filter, channel, height, and width, respectively. 
Our proposed objective function for regularization can be formulated as
\begin{equation}
\label{eqn:objective function}
E(\mathbf{W}) = L(\mathbf{W}) + \frac{\lambda}{2}R(\mathbf{W}) + \sum_{l=1}^L \sum_{g=1}^{G^{(l)}} \frac{\lambda_g^{(l)}}{2} R(\mathbf{W}^{(l)}_g),
\end{equation}
where~$\mathbf{W}$ denotes the collection of all weights in the CNN; $L(\mathbf{W})$ is the loss function for prediction; $R(\mathbf{W})$ is non-structured regularization on every weight, \emph{i.e.}, weight decay in this paper; $R(\mathbf{W}^{(l)}_g)$ is the structured sparsity regularization term on group $g$ of layer $l$ and $G^{(l)}$ is the number of weight groups in layer $l$. 
In~\cite{VadLem16,wen2016learning}, the authors use the same~$\lambda_g$ for all groups and adopt Group LASSO~\cite{Yuan2006Model} for~$R(\mathbf{W}^{(l)}_g)$. 
In this work, since we emphasize the key problem of group regularization lies in the regularization factor rather than the exact regularization form, we use the most common regularization form weight decay, for~$R(\mathbf{W}^{(l)}_g)$, but we vary the regularization factors~$\lambda_g$ for different weight groups and at different iterations.

\begin{figure*}
   \centering
   \includegraphics[width=0.9\textwidth, height=0.18\textheight]{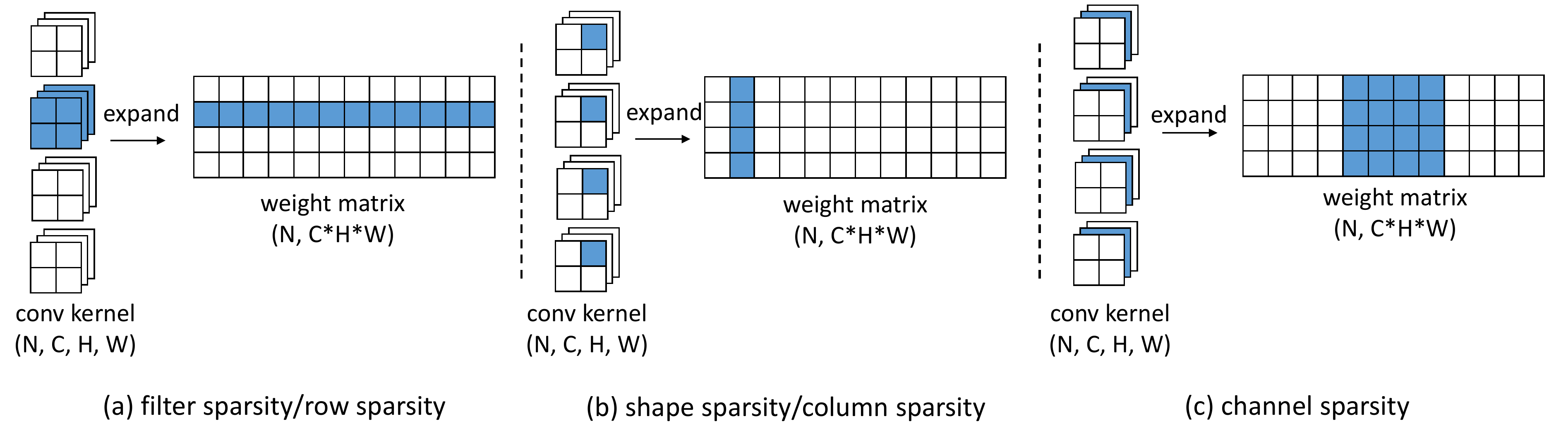}
   \caption{The im2col implementation of CNN is to expand tensors into matrices, so that convolution is transformed to matrix multiplication. The blue squares represent pruned weights.  (a) Pruning a row of weight matrix is equivalent to pruning a filter in convolutional kernel. (b) Pruning a column of weight matrix is equivalent to pruning all the weights at the same position in different filters. (c) Pruning a channel is  equivalent to pruning several adjacent columns in the expanded weight matrix. }
\label{fig:sparsity_structure}
\end{figure*}

The final learned sparsity structure depends on the way of splitting groups of~$\mathbf{W}^{(l)}$. 
In the \verb+im2col+ implementation of convolution~\cite{ChePurSim06,CheWooVan14}, there are normally three kinds of sparsity groups in a layer, \emph{i.e.}, filter-wise (\emph{a.k.a.} row sparsity), channel-wise, and shape-wise (\emph{a.k.a.} column sparsity) sparsity~\cite{wen2016learning}, as shown in Figure \ref{fig:sparsity_structure}.
Practical acceleration can be achieved by removing the zero rows and columns when the weight and feature tensors are lowered into matrices~\cite{wen2016learning,CheWooVan14}, which is easy to implement on popular deep learning platforms such as Caffe, TensorFlow, PyTorch, \emph{etc}.

\subsection{Theoretical Analysis}
For the proposed method, the regularization factor~$\lambda_g$ is differently assigned for different weight groups. 
It can be seen that by slightly augmenting~$\lambda_g$ of a weight group, then training the network through back-propagation until the objective function reaches the local minimum, the~$L_2$-norm of that weight group will also decrease. 
This idea is formally summarized by the following theorem.
\begin{theorem}
\label{theorem:main}
Consider the objective function
\begin{equation}
E(\lambda,\omega) = L(\omega) + \frac{\lambda}{2} \omega^2,
\end{equation}
if there exists a tuple~$(\lambda_0,\omega_0)$ which satisfies the following three properties:
\begin{enumerate}
\item \label{property 1} $\lambda_0>0$,
\item \label{property 2} $L(\omega)$ has the second derivative at~$\omega_0$,
\item \label{property 3} $\omega_0$ is the local minimum of function~$Y_{\lambda_0}(\omega) = E(\lambda_0,\omega)$,
\end{enumerate}
then there exists an~$\epsilon>0$ that for any~$\lambda_1 \in (\lambda_0, \lambda_0+\epsilon)$, we can find an~$\omega_1$ which satisfies:
\begin{enumerate}
\item $\omega_1$ is the local minimum of function~$Y_{\lambda_1}(\omega) = E(\lambda_1,\omega)$,
\item $|\omega_1| < |\omega_0|$.
\end{enumerate}
\end{theorem}

This theorem is proved in Appendix A, which indicates that we can slightly increase the regularization factor to compress the magnitude of weights to zero. By Equation (\ref{eqn:dlambda0}), the magnitude of weights will be more compressed if the regularization factor increases more.

\subsection{Method Description}
Theorem~\ref{theorem:main} guarantees that we can modify the~$L_1$-norm of weight groups by increasing or decreasing their corresponding regularization factors. 
Thus, we can assign different regularization factors to weight groups based on their importance to the network. 
In this paper, $L_1$-norm is used as the importance criterion for its simplicity. 
Note that our method can be easily generalized to other criteria such as~$L_2$-norm, APoZ~\cite{hu2016trimming}, and Taylor expansions~\cite{lecun1990optimal,MolTyrKar17}.

Normalization of importance criteria is necessary because the values of $L_1$-norms have huge variation across different networks, layers, and weight groups. 
The normalization in our method is based on the ranks of weight groups in the same layer. 
The advantages of rank-based normalization lies in two parts: (1) Compared to other normalization methods like max/min normalization, the range of ranks is \emph{fixed} from~$0$ to~$N_g-1$, where~$N_g$ is the total number of weight groups in the layer; (2) For the pruning task, we need to set a pruning ratio~$R$ to each layer, say, $R=0.6$ means that we need to prune~$60\%$ of weight groups which are ranked the lowest when pruning is finished. 
Normalization by ranks makes the pruning process controllable since it is directly towards the goal of pruning.

Specifically, we sort weight groups by their $L_1$-norms in ascending order. 
Meanwhile, to mitigate the oscillation of ranks in one training iteration, we average the rank of each group through training iterations. 
For a weight group, its average rank through~$N$ iterations is defined as
\begin{equation}
	\overline{r}_{avg} = \frac{1}{N}\sum_{n=1}^{N} r_n.
\label{eqn:ave_hrank}
\end{equation}
Here~$r_n$ is the rank of the~$n$th iteration. The final average rank~$\overline{r}$ is obtained by sorting~$\overline{r}_{avg}$ of different weight groups in ascending order, making its range from~$0$ to~$N_g-1$.

Our aim is to assign an increment~$\Delta \lambda_g$ to each weight group, so that its regularization factor~$\lambda_g$ is gradually updated through the pruning process:
\begin{equation}
	\lambda_g^{(new)} = \lambda_g+\Delta \lambda_g.
\label{eqn:changing lambda}
\end{equation}

Following the above idea, $\Delta \lambda_g$  of each group is assigned by its average rank~$\overline{r}$ with a proposed piecewise linear function,
\begin{equation}
\Delta\lambda_g(\overline{r}) = \left\{
		\begin{aligned}
			& -\frac{A}{RN_g}\overline{r}+A            &  \text{\emph{ if} }  \overline{r} \leq RN_g;\\
			&-\frac{A}{N_g(1-R)-1}(\overline{r}-RN_g)  &  \text{\emph{ if} } \overline{r} >  RN_g.
		\end{aligned}
    \right.
\label{eqn:punish_function}
\end{equation}

\begin{figure}
   \centering
   \includegraphics[width=0.8\linewidth]{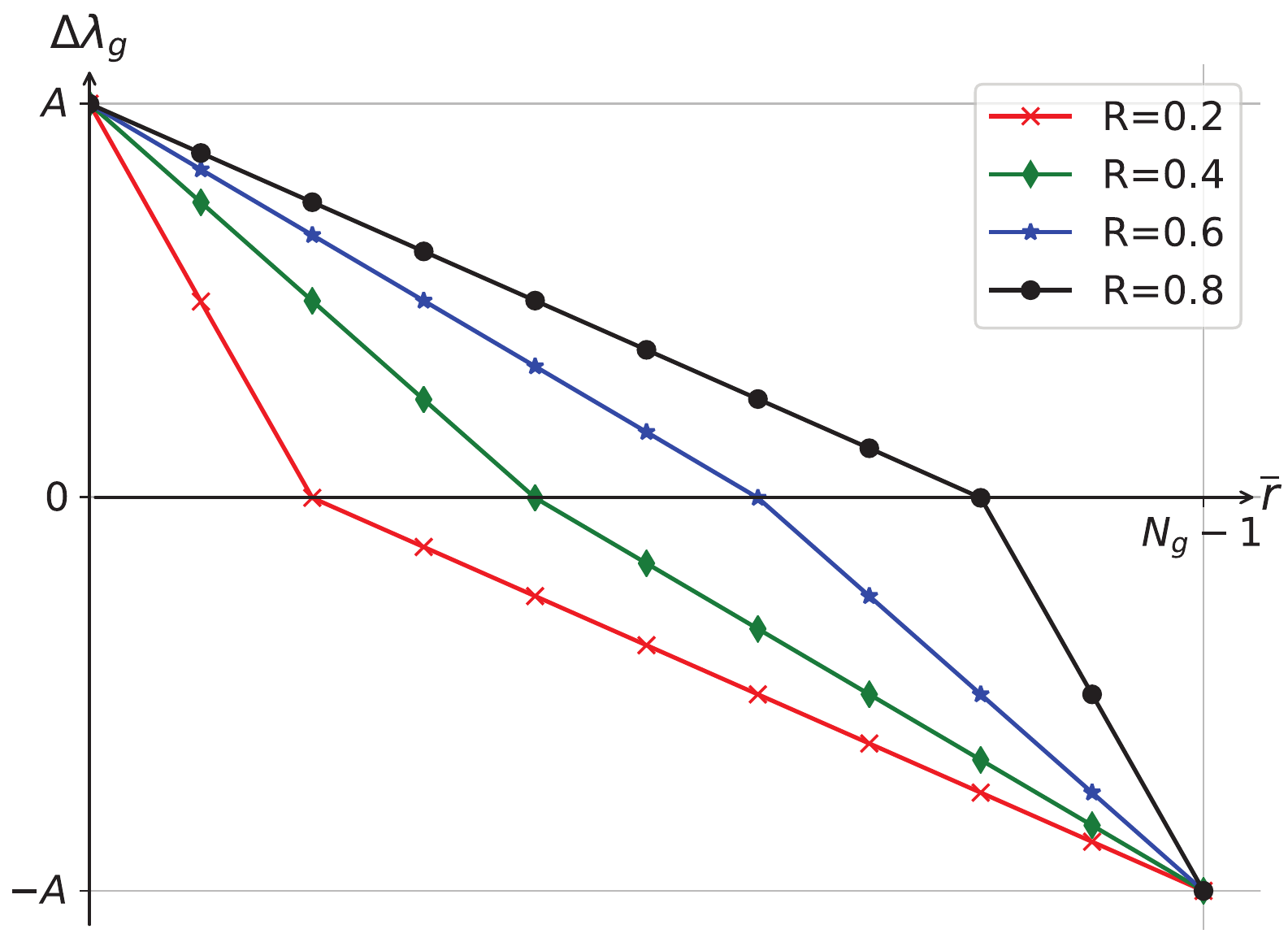}
   \caption{The functional of~$\Delta \lambda_g(\overline{r})$, as defined in Equation (\ref{eqn:punish_function}).}
\label{fig:punish_function}
\end{figure}

Figure \ref{fig:punish_function} depicts~$\Delta \lambda_g(\overline{r})$, where it can be seen that for weight groups whose~$L_1$-norms are small, \emph{i.e.}, the average ranks less than~$RN_g$, we need to increase their regularization factors to further decrease their~$L_1$-norms; and those with greater~$L_1$-norms and rank above~$RN_g$, we need to decrease regularization factors to further increase their~$L_1$-norms.   
After obtaining~$\Delta \lambda_g$ and~$\lambda_g^{(new)}$ by Equation (\ref{eqn:punish_function}) and~(\ref{eqn:changing lambda}), we threshold it by zero to prevent negative values of regularization:
\begin{equation}
\lambda_g^{(new)} = \max(\lambda_g+\Delta \lambda_g,0).
\label{eqn:lambda update}
\end{equation}

After updating~$\lambda_g$, the weights of CNN are trained through back-propagation deduced from Equation (\ref{eqn:objective function}). 
If a weight is smaller than some pre-defined threshold, it will be eliminated from the network permanently.
After training the weights for several iterations, we recalculate~$\lambda_g$ and the training process continues until convergence.   
Since we decrease the~$L_1$-norms of weight groups whose ranks are less than~$RN_g$ and increase the~$L_1$-norms of weight groups whose ranks are greater than~$RN_g$, there should be exactly $RN_g$ pruned weight groups at the convergence point.   
In Equation (\ref{eqn:punish_function}), $A$ is a hyper-parameter to control the speed of convergence.   
Greater value of~$A$ results in faster convergence.   
To this end, we can summarize the proposed algorithm in Algorithm~\ref{alg:VR}.

\begin{algorithm}[t]
\caption{The Proposed Algorithm}
\begin{algorithmic}[1]
\State Input the training set~$D$, the original pre-trained model~$\Omega$, the non-structural regularization factor~$\lambda$, pruning threshold~$\epsilon = 10^{-5}$, and target pruning ratio~$R_l$ for each convolutional layer $l$.
\State Initialize~$\lambda_g =0$ for each weight group.
\State Initialize the iteration number~$i=0$.
\Repeat
    \For{each weight group in each layer}
      \State Obtain~$\overline{r}$ of by sorting $\overline{r}_{avg}$ as Equation (\ref{eqn:ave_hrank}).
      \State Update~$\lambda_g$ by Equation (\ref{eqn:punish_function}) and (\ref{eqn:lambda update}).
    \EndFor
    \State Update weights in each weight groups through back-propagation.
    \For{each weight group in each layer}
      \If {the~$L_1$-norm of this group is less than~$\epsilon$}
         \State prune this weight group permanently.
      \EndIf
    \EndFor
    \State $i=i+1$.
\Until {The ratio of pruned weight groups of each layer reaches~$R_l$.}
\State Retrain the pruned CNN for several epochs.
\State Output the pruned CNN model~$\Omega'$.
\end{algorithmic}
\label{alg:VR}
\end{algorithm}

A nice characteristic of the proposed method is that it can automatically adjust searching steps without any knowledge about the property of the objective function itself. 
Specifically, by Equation (\ref{eqn:dlambda0}), the increment of weight~$d\omega$ is reversely proportional to the second derivative of the objective function~$L(\omega)$, which makes the modification of~$\omega$ slower when the objective function~$L(\omega)$ reaches steeper areas even without knowing the exact form of~$L(\omega)$.
This is a nice property to alleviate the difficulty of optimization and make refined searching for possibly better local minimums.
We believe that the good performance of the proposed method is partly attributed to the automatic adjustment of the searching steps.

\section{Experiments}
We firstly analyze the difference between the proposed incremental regularization and existing constant regularization, with ConvNet on CIFAR-10 dataset~\cite{KriHin09}. 
Then we evaluate the proposed method with deep CNNs on the large-scale ImageNet dataset. 
All of our experiments are conducted with Caffe~\cite{JiaSheDonEtAl14}.
The codes and trained models are available to public: \href{https://github.com/mingsun-tse/caffe\_increg}{https://github.com/mingsun-tse/caffe\_increg}.

The only hyper-parameter in our method is $A$, which is set to half of the original weight decay unless specially mentioned. 
The other hyper-parameters (such as weight decay, momentum, \emph{etc}.) are kept unchanged as their original values. 
Since this work focuses on CNN acceleration rather than compression, we only prune weights in the \emph{convolutional} layers considering the little computation proportion in fully-connected layers. 
For all experiments, speedup is calculated by GFLOPs reduction.

\subsection{Analysis with ConvNet on CIFAR-10}
We compare IncReg with two kinds of regularization-based structured pruning methods. 
(1) The first is Group LASSO~\cite{Yuan2006Model,VadLem16,wen2016learning}, where the regularization factor is \emph{uniform} for all weight groups and \emph{constant} during the whole pruning process. 
SSL~\cite{wen2016learning} is chosen as representative. 
(2) The second is \emph{auto-balanced} regularization scheme, proposed in AFP~\cite{DinDinHanTan18}, where regularization factors are differentiated by the importance of weight groups, \emph{i.e.}, more important weight groups are punished less, partly similar to our work, but they are still \emph{constant} during pruning. 
We implemented AFP with Caffe~\cite{JiaSheDonEtAl14}.

The test network is ConvNet, a small CNN adapted from AlexNet~\cite{KriSutHin12} with three convolutional layers and one fully-connected layers.  
CIFAR-10 is a $10$-class dataset of $60k$ tiny images, among which $45k$ images are used for training, $5k$ for validation and the other $10k$ for testing.  
We first trained a baseline model with testing accuracy $81.5\%$.  
Then the three kinds of regularization schemes are applied to learn structured sparsity, where both row and column sparsity are explored.

Experimental results are shown in Table \ref{tab:compare_ssl_ours}.  
We can see that IncReg consistently achieves higher speedups and accuracies than the constant regularization schemes (SSL and AFP).
Notably, even though AFP achieves similar performance as our method under relatively small speedup (about $4.5\times$), when the speedup ratio is large (about $8\times\sim10\times$), our method outperforms AFP by a large margin.

\begin{figure}[!h]
   \centering
   \includegraphics[width=0.48\textwidth]{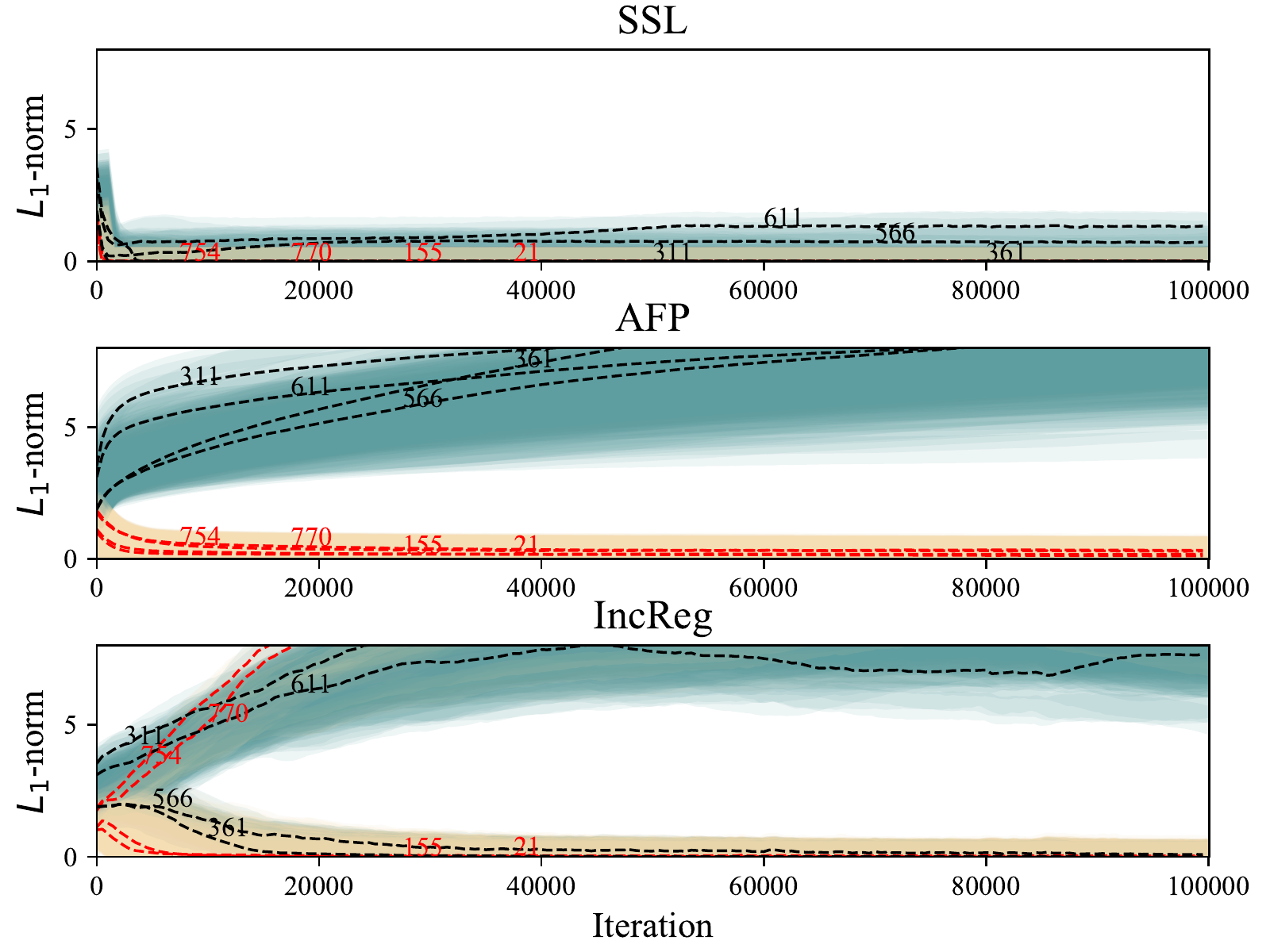}
   \caption{Comparison of the pruning process of SSL, AFP, and IncReg. Each line illustrates the $L_1$-norm of a column in the conv2 layer ($800$ columns in total) of ConvNet. Based on their ranks by $L_1$-norm at iter $0$, columns are divided into two groups, of which the smallest $800R$ columns ("unimportant columns") are colored by light yellow, while the other $800(1-R)$ largest columns ("important columns") are color by light green. $R=0.78$ for this plot. Eight column examples (the dashed lines) are depicted with their column numbers shown next to the lines). Among them, black lines denote the "important columns"; red lines denote the "unimportant columns" (\emph{Best seen in color}).}
   \label{fig:compare_fixed_varying}
\end{figure}

To compare the difference among the three schemes more vividly, the pruning processes (scenario: column pruning, large speedup) of SSL, AFP, and IncReg are illustrated in Figure \ref{fig:compare_fixed_varying}, where we depict how the~$L_1$-norms of columns in the \verb+conv2+ layer change during the whole the training process.
The important and unimportant columns (measured by their $L_1$-norms at iter $0$) are colored by light green and light yellow respectively.
Due to the large number of them ($800$ columns in total, \emph{i.e.}, $800$ lines), the exact one line of them can hardly be made out, so we also depict eight columns with the dashed line as representatives.
The number next to each dashed line is its column number.
From the figure, we can see although SSL manages to drive the unimportant weight groups towards zero, yet many important weight groups are also suppressed unnecessarily, which probably lead to irreversible damage to the energy or expressiveness of weights, answering for the under-performance of SSL (Table \ref{tab:compare_ssl_ours}).  
This problem is noticed in AFP~\cite{DinDinHanTan18}, so they propose an \emph{explicit} auto-balanced regularization scheme to maintain the total energy of weights, where the unimportant weight groups are punished (with positive regularization), meanwhile, the important weight groups are stimulated (with negative regularization).  
Notably, as is shown in Figure \ref{fig:compare_fixed_varying}, although our method only imposes positive regularization on the unimportant weight groups, the important weight groups increase their $L_1$-norms \emph{automatically}, \emph{i.e.}, the network actually \emph{learns by itself} to balance the weight energy without any explicit stimulation.
Therefore, the incremental way of regularization is \emph{naturally} beneficial for the network to adapt during pruning.  
Besides, the slope of $L_1$-norm trajectories using IncReg is less steep than that using AFP, which means the pruning process of IncReg is more gentle.
This gentleness is valuable for the network to transfer its fragile expressiveness to the remainder of the network, especially when pruning a large proportion of parameters, which can explain why IncReg is more robust than AFP under the scenario of $8\times\sim10\times$ speedup (Table \ref{tab:compare_ssl_ours}).

Another notable point in Figure \ref{fig:compare_fixed_varying} is that, for Column $754$ and $770$ of IncReg, although they are regarded as unimportant at iter $0$, they actually become important later (note that their $L_1$-norms rise up). 
The similar story also happens to Column $361$ and Column $566$ except that they become unimportant while they are viewed as important at the beginning. 
This phenomenon shows that IncReg does not determine the importance of a weight group once for all. 
Instead, it collects evaluations over the whole training process and has the ability to correct the importance misjudgments.
This flexibility is enabled by the proposed incremental regularization scheme, which is however not shared by the large constant regularization schemes (SSL and AFP).

Actually, there is another bonus from IncReg.  
We varies the only hyper-parameter $A$ in our method to see how it influences the performance.  
The accuracies are shown in Table \ref{tab:sensitivity_of_A}, where even if the range of $A$ varies by an order of magnitude (from $10^{-4}$ to $10^{-3}$), the performance only has minute changes ($0.1\%\sim0.2\%$). 
Namely, the performance of IncReg is especially robust to the change of hyper-parameter, which can mean a lot by liberating practitioners from hard hyper-parameter tuning.
Note that this robustness cannot translate to common constant regularizations like $L_2$ or $L_1$ used in SSL and AFP, because their values are usually much larger and uniformly applied to all the weights.

\begin{table}
    \centering
    \begin{tabular}{l p{1cm}<{\centering} p{1cm}<{\centering} p{1cm}<{\centering} p{1cm}<{\centering}}
       \toprule
         \multirow{2}*{Method}  & \multicolumn{2}{c}{Row pruning} & \multicolumn{2}{c}{Column pruning} \\
         		   \cline{2-5}
                   & speedup & accuracy & speedup & accuracy \\
         \hline
         SSL               & $3.6 \times$    & $77.3$          & $3.1\times$    & $78.6$  \\
         AFP~(our impl.)   & $4.1\times$     & $77.7$           & $4.5\times$          & $81.0$ \\
         IncReg            & $\mathbf{4.1\times}$           & $\mathbf{79.2}$ & $\mathbf{4.6\times}$   & $\mathbf{81.2}$ \\
         \hline
         SSL               & $9.7\times$      & $73.0$              & $10.0\times$    & $75.2$ \\
         AFP~(our impl.)   & $9.9\times$      & $73.4$              & $8.4\times$          & $77.5$ \\
         IncReg            & $\mathbf{9.9\times}$      & $\mathbf{76.0}$              & $\mathbf{10.0\times}$    & $\mathbf{78.7}$ \\                  
       \bottomrule
    \end{tabular}
    \caption{Comparison of varying and fixed regularization with ConvNet on CIFAR-10. The baseline testing accuracy is $81.5\%$.}
    \label{tab:compare_ssl_ours}
\end{table}

\begin{table}
    \centering
    \begin{tabular}{l p{0.47cm}<{\centering} p{0.47cm}<{\centering} p{0.47cm}<{\centering} p{0.47cm}<{\centering} p{0.47cm}<{\centering} p{0.47cm}<{\centering} p{0.47cm}<{\centering}}
       \toprule
         $A$ ($10^{-4}$)          & $1$         & $2.5$ & $5$        & $7.5$ & $10$ & $20$ & $40$  \\
         \midrule
         Accuracy (\%) & $\mathbf{81.4}$   & $81.3$     & $\mathbf{81.4}$  & $81.2$     & $81.3$   & $81.1$   & $80.9$\\
       \bottomrule
    \end{tabular}
     \caption{Accuracy comparison w.r.t. different $A$'s with ConvNet on CIFAR-10 under $4\times$ speedup. Default setting is $A=0.0005$. For reference, the original weight decay is $0.004$.}
    \label{tab:sensitivity_of_A} 
 \end{table}

In addition, from Table \ref{tab:compare_ssl_ours} we find that under similar speedup ratios, column pruning is better than row pruning in accuracy.  
It is probably because a row typically consists of much more weights than a column does, therefore row pruning causes more severe side-effects to the expressiveness of network.  
Given row and column sparsity can achieve similar practical speedup (see Table \ref{tab:result_vgg16}), in the following experiments, we only choose \emph{column} as our sparsity group to obtain better performance.
We further compare the proposed method with another related pruning methods. The results are shown in Table \ref{tab:acc_convnet_cifar10}.  
Under different speedup ratios, our method consistently outperforms other pruning methods, and for small speedup ratios (like $2\times$) our method even improves the performance, and by more than the others do.

\begin{table}
    \newcommand{\tabincell}[2]{\begin{tabular}{@{}#1@{}}#2\end{tabular}}
    \centering
    \begin{tabular}{lccc}
       \toprule
         \multirow{2}*{Method}  &  \multicolumn{3}{c}{Increased err. (\%)} \\
         \cline{2-4}
			  & $2\times$ & $4\times$ & $6\times$\\
         \midrule
         TP~\cite{MolTyrKar17}~(our impl.)   &  $1.0$ & $3.4$ & $5.4$  \\
         FP~\cite{LiKadDurEtAl17}~(our impl.)   &  $1.6$ & $3.6$ & $5.0$  \\
         SPP~\cite{wang2017structured}  & -$0.2$ & $0.3$ & $1.2$ \\
         Ours & -$\mathbf{0.5}$ & $\mathbf{0.1}$ & $\mathbf{1.0}$ \\
       \bottomrule
    \end{tabular}
     \caption{The increased error of different pruning methods when accelerating ConvNet on CIFAR-10. The baseline testing accuracy is $81.5\%$. Minus means the testing accuracy is improved.}
    \label{tab:acc_convnet_cifar10}
 \end{table}

\subsection{ResNet-56 on CIFAR-10}
We further evaluate our method on a network with different architecture paradigm. 
Different from ConvNet, ResNet-56~\cite{HeZhaRenSun16} is a multi-branch residual network, which is much deeper and more compact. 
We train our own baseline model with batch size $128$ and base learning rate $0.1$, following the instructions in~\cite{HeZhaRenSun16} (except that we do not use the zero-padding data augmentation). 
The testing accuracy on CIFAR-10 is $93.0\%$, identity with the original performance~\cite{HeZhaRenSun16}. 
ResNet-56 has $57$ convolutional layers, followed by a global average pooling and fully-connected Softmax classification layer.
%
Uniform pruning ratios are adopted for all $55$ convolutional layers, with the two $1\times1$ projection shortcuts spared given their ignorable computation. 
After training with batch size $64$ and fixed learning rate $0.025$, the pruning is done, followed by a retraining process with original batch size $128$ and step-decreased learning rate. 

Results are shown in Table \ref{tab:result_resnet56}, where IncReg consistently outperforms the other methods \emph{by a large margin} under different speedup ratios. 
Especially, under relatively small speedup ($2\times$), none of methods but IncReg can even improve the performance. 
Besides, note that even though AFP~\cite{DinDinHanTan18} uses a better baseline model (with accuracy $93.9\%$ vs. ours $93.0\%$), when pruned with a large speedup ratio (\emph{e.g.}, $3.4\times$), their performance degrades quickly (by $3.4\%$), while our method still degrades little (by only $0.9\%$ and with more speedup). 
We argue that this robustness is attributed to the \emph{incremental} nature of our method discussed above.

\begin{table}
    \centering
    \begin{tabular}{p{2.5cm}  p{1.5cm}<{\centering} p{2.0cm}<{\centering}}
       \toprule
         Method & Speedup & Increased err.(\%)\\
         \midrule
         FP~\cite{LiKadDurEtAl17} (CP's impl.)                   & $2.0\times$   & $1.3$ \\
         CP~\cite{he2017channel}                                 & $2.0\times$   & $1.0$ \\
         SPP~\cite{wang2017structured}            & $2.0\times$   & $0.1$ \\
         AMC~\cite{he2018amc}                                 & $2.0\times$   & $0.9$ \\    
         Ours~($R=0.500$)                                        & $\mathbf{2.1\times}$   & -$\mathbf{0.3}$ \\
         \hline
         AFP-1~\cite{DinDinHanTan18}                             & $2.6\times$   & $1.0$ \\
         Ours~($R=0.608$)                                        & $\mathbf{3.1\times}$   & $\mathbf{0.4}$ \\
         \hline
         AFP-2~\cite{DinDinHanTan18}                             & $3.4\times$   & $3.4$ \\
         Ours~($R=0.667$)                                      & $\mathbf{4.0\times}$  & $\mathbf{0.9}$ \\
       \bottomrule
    \end{tabular}
    \caption{Acceleration of ResNet-56 on CIFAR-10. The baseline accuracy of the original network is~$93.0\%$. For reference, the baseline model of AFP is $93.9\%$ using TensorFlow implementation.}
    \label{tab:result_resnet56}
\end{table}

\subsection{AlexNet on ImageNet}
\label{sec:exp_alex}
We apply the proposed method to AlexNet~\cite{KriSutHin12}, which is composed of $5$ convolutional layers and $3$ fully-connected layers. 
We download an open caffemodel from Caffe model zoo as our pre-trained model. 
The baseline single view top-5 accuracy on ImageNet 2012 validation dataset is $80.0\%$. 
All images are rescaled to size $256\times256$, then a $227\times227$ patch is randomly cropped from each scaled image and randomly mirrored for data augmentation. 
For testing, the $227\times227$ patches are cropped from the center of the scaled images.

Intuitively, different layers have different sensitivity to pruning, but there are few theories to quantify the redundancy of different layers in deep neural networks.
Most pruning methods empirically set pruning ratios for different layers~\cite{LiKadDurEtAl17,MolTyrKar17,he2017channel,wang2017structured}. 
Aligned with these works, we empirically set the proportion of remaining columns of the $5$ convolutional layers as~$1:1:1:1.5:1.5$.
We train the baseline model with batch size $64$, fixed learning rate $0.0005$, for about $7$ epochs before reaching the target pruning ratios. 
Then the pruned model is retrained with batch size $256$ to regain accuracy.

Experimental results are shown in Table \ref{tab:result_alexnet}. 
The proposed method and SPP are consistently better than the other three methods, and the proposed method is slightly better than SPP on average. 
With $4\times$ speedup, our method can even improve the accuracy while SPP degrades the accuracy in this situation.

\begin{table}[ht]
    \centering
    \begin{tabular}{lccc}
       \toprule
         \multirow{2}*{Method}  &  \multicolumn{3}{c}{Increased err. (\%)} \\
         \cline{2-4}
                  & $2\times$ & $4\times$ & $5\times$ \\
         \hline
         TP~\cite{MolTyrKar17}       & $3.9$  & $9.2$   & $13.9$  \\
         FP~\cite{LiKadDurEtAl17} (our impl.)      & $0.6$  & $4.1$   & $4.7$   \\
         SSL~\cite{wen2016learning}      & $1.3$  & $4.3$   & $5.3$   \\
         SPP~\cite{wang2017structured}      & -$\mathbf{0.7}$ & $0.3$ & $0.9$   \\
         Ours     & -$\mathbf{0.7}$ & -$\mathbf{0.2}$ &  $\mathbf{0.8}$ \\
       \bottomrule
    \end{tabular}
     \caption{Acceleration of AlexNet on ImageNet. The baseline top-5 accuracy of the original network is $80.0\%$.}
    \label{tab:result_alexnet}
\end{table}

\begin{table*}[!h]
    \centering
    \begin{tabular}{l p{1.5cm}<{\centering} p{1.5cm}<{\centering} p{1.5cm}<{\centering} p{1.5cm}<{\centering} p{1.5cm}<{\centering} p{1.5cm}<{\centering}}
       \toprule
         \multirow{2}*{Method}    & \multicolumn{3}{c}{CPU time (baseline: 1815 ms)}          & \multicolumn{3}{c}{GPU time (baseline: 5.159 ms)} \\
                                  \cline{2-7}
                                  & $2\times$ & $4\times$ & $5\times$                         & $2\times$ & $4\times$ & $5\times$ \\
         \hline         
         CP~\cite{he2017channel}  & $826 (2.2\times)$ & $500 (3.6\times)$ & $449 (4.0\times)$ & $3.206 (1.6\times)$ & $2.202 (2.3\times)$ & $2.034 (2.5\times)$ \\
         Ours                     & $861 (2.1\times)$ & $469 (3.9\times)$ & $409 (4.4\times)$ & $3.225 (1.6\times)$ & $2.068 (2.5\times)$ & $1.991 (2.6\times)$ \\
       \bottomrule
    \end{tabular}
    \caption{Inference time of convolutional layers of CP and our method on VGG-16. Evaluation is carried out with batch size $10$ and averaged by $50$ runs on $224\times224$ RGB images. CPU: Intel Xeon(R) E5-2620 v4 @ 2.10GHz, single thread; GPU: GeForce GTX 1080Ti, without cuDNN~\cite{CheWooVan14}. Open source models of CP are used for this evaluation.}
    \label{tab:actual_speedup_vgg16}
\end{table*}

\subsection{VGG-16 on ImageNet}
\label{sec:exp_vgg}
We further demonstrate our method on VGG-16~\cite{Simonyan2014Very}, which has $13$ convolutional layers and $3$ fully-connected layers. 
We download the open caffemodel as our pre-trained model, whose single-view top-5 accuracy on ImageNet 2012 validation dataset is $89.6\%$. 
Data augmentation is like the part of AlexNet experiment (Section~\ref{sec:exp_alex}) with input size $224\times224$.

Previous works~\cite{he2017channel,wang2017structured} find that lower layers are more redundant in VGG-16. 
Therefore, the proportion of remaining ratios of low layers (\verb+conv1_x+ to \verb+conv3_x+), middle layers (\verb+conv4_x+) and high layers (\verb+conv5_x+) are set to $1:1.5:2$, the same as~\cite{wang2017structured} for easy comparison. 
The first and last convolutional layer (\verb+conv1_1+ and \verb+conv5_3+) are not pruned because of their little computation. 
Like the experiment of AlexNet, we first train the baseline model with batch size $64$ and fixed learning rate $0.0005$. 
Pruning is finished after around~$6$ epochs. 
Then the network is fine-tuned with batch size $256$ to regain accuracy.

Experimental results are shown in Table \ref{tab:result_vgg16}. 
Our method is slightly better than CP and SPP, and outperforms TP and FP by a significant margin. 
Recently, AMC~\cite{he2018amc} used reinforcement learning to search for the best combination of pruning ratios in different layers. 
Notably, even with optimized pruning ratio combination (at much more computation cost than ours), their method is only slightly better than ours.
And since we use the same pruning ratios as SPP does, the only explanation for the performance improvement should be a better pruning process itself, demonstrating the effectiveness of the proposed incremental regularization scheme. 
On top of GFLOPs reduction, we also evaluate the practical speedup in terms of inference time reduction, compared with CP on both CPU and GPU platform.
The results are shown in Table \ref{tab:actual_speedup_vgg16}.

\begin{table}[!h]
    \centering
    \begin{tabular}{l p{1cm}<{\centering} p{1cm}<{\centering} p{1cm}<{\centering}}
       \toprule
         \multirow{2}*{Method}  &  \multicolumn{3}{c}{Increased err. (\%)} \\
         \cline{2-4}
         & $2\times$ & $4\times$ & $5\times$ \\
         \hline
         TP~\cite{MolTyrKar17}              & $-$ & $4.8$ & $-$   \\
         FP~\cite{LiKadDurEtAl17} (CP's impl.) & $0.8$ & $8.6$ & $14.6$ \\
         CP~\cite{he2017channel}              & $\mathbf{0}$ & $1.0$ & $1.7$ \\
         SPP~\cite{wang2017structured}             & $\mathbf{0}$  & $\mathbf{0.8}$  & $2.0$ \\
         AMC~\cite{he2018amc}             & $-$           & $-$             & $\mathbf{1.4}$ \\
         Ours            & $\mathbf{0}$  & $\mathbf{0.8}$   & $1.5$ \\
       \bottomrule
    \end{tabular}
     \caption{Acceleration of VGG-16 on ImageNet. The values are increased single-view top-5 error on ImageNet. The baseline top-5 accuracy of the original network is $89.6\%$.}
    \label{tab:result_vgg16}
\end{table}

\vspace{-1.5em}
\begin{table}[!h]
    \centering
    \begin{tabular}{l p{2.5cm}<{\centering}}
       \toprule
         Method  & Increased err. (\%) \\
         \midrule
         CP~\cite{he2017channel}  & $1.4$ \\
         SPP~\cite{wang2017structured} & $0.8$ \\
         Ours & $\mathbf{0.1}$ \\
       \bottomrule
    \end{tabular}
     \caption{Acceleration of ResNet-50 on ImageNet. The baseline top-5 accuracy of the original network is $91.2\%$.}
    \label{tab:result_resnet50}
 \end{table}

\subsection{ResNet-50 on ImageNet}
Unlike AlexNet and VGG-16, ResNet-50~\cite{HeZhaRenSun16} is a multi-branch compact deep neural network, which has $53$ convolutional layers and one fully-connected classification layer. 
Open pre-trained caffemodel is adopted as baseline, whose single view top-5 accuracy on ImageNet 2012 validation dataset is $91.2\%$. 
The images are augmented in the same way of the VGG-16 experiment. 
For simplicity, we adopt the same pruning ratio~$0.4$ for all the $53$ convolutional layers. 
The training settings are similar to that of VGG-16. 
The pruning process stops after less than $2$ epochs before retraining.

From Table \ref{tab:result_resnet50}, our method achieves $2\times$ speedup with only $0.1\%$ performance loss, significantly better than CP and SPP. 
It should be mentioned for fair comparison that CP used a better baseline model ($92.2\%$ vs. ours $91.2\%$), but still, their absolute performance is worse than ours. 
This might be because the feature reconstruction scheme in CP is not very effective for the compact residual networks, while a possibly better way around is to let the network learn to recover by itself. 
Considering the compact nature of residual networks, intuitively, an incremental way of pruning is more gentle and thus more beneficial to its recovery. 
Although in SPP, the network learns to recover, but given the probabilistic way of training, it is probably too dynamic for the compact residual network. 
In comparison, our method addresses the dynamics of training in a more gentle way, thus delivering a better result.

\section{Conclusion}
We propose a new structured pruning method based on an incremental regularization scheme with a relatively sound theoretical basis, which helps CNN to effectively transfer its expressiveness to the remainder of network during pruning by adjusting the regularization factors gradually.
Theoretical analysis guarantees the convergence of our method, and its effectiveness is proved by extensive experimental comparisons with state-of-the-art methods on popular CNN architectures.

One less principled aspect of our method is that the assignment of regularization increment is done by the pre-defined function rather than via optimization.
If the second derivatives are available, the assignment can be formulated as a more disciplined optimization problem, which falls into our future work. 
Another line of future work includes generalizing the incremental regularization scheme to other regularization-based pruning methods (\emph{e.g.}, network slimming~\cite{liu2017learning}) and other network architectures (\emph{e.g.}, recurrent neural networks).

%

\section*{Appendix A --  Proof of Theorem~\ref{theorem:main}}
\noindent \textbf{Proof:} For a given~$\lambda>0$, the~$\omega$ which is the local minimum of the function~$Y_{\lambda}(\omega) = E(\lambda,\omega)$ should satisfy~$\frac{dY_{\lambda}(\omega)}{d\omega}=0$, which gives $\lambda = -\frac{L'(\omega)}{\omega}$. 
Then we can obtain the derivative of~$\lambda$ w.r.t. $\omega$,
\begin{equation}
\label{eqn:dlambda}
	\frac{d\lambda}{d\omega} = \frac{L'(\omega)-\omega L''(\omega)}{\omega^2}.
\end{equation}

Since~$\omega_0$ is the local minimum of the function~$Y_{\lambda_0}(\omega) = E(\lambda_0,\omega)$, it should satisfy
\begin{equation}
\left. \frac{dY_{\lambda_0}(\omega)}{d\omega} \right|_{\omega_0}=0 \textrm{ and } \left. \frac{d^2Y_{\lambda_0}(\omega)}{d\omega^2} \right|_{\omega_0}>0, \nonumber
\end{equation}
which yields
\begin{eqnarray}
\label{eqn:first}
L'(\omega_0)+\lambda_0 \omega_0&=&0, \\
\label{eqn:second}
L''(\omega_0)+\lambda_0 &>&0.
\end{eqnarray}

If we take~Equation (\ref{eqn:first}) into~(\ref{eqn:dlambda}), we can obtain
\begin{equation}
\label{eqn:dlambda0}
\left.\frac{d\lambda}{d\omega} \right|_{(\lambda_0,\omega_0)}=-\frac{L''(\omega_0)+\lambda_0}{\omega_0}.
\end{equation}

Then by taking~Equation (\ref{eqn:second}) into~(\ref{eqn:dlambda0}), we can conclude
\begin{equation}
 \left\{
		\begin{aligned}
	  		      & \left.\frac{d\lambda}{d\omega} \right|_{(\lambda_0,\omega_0)}<0,          &  \text{ if }  \omega_0>0;  \\
	  		       & \left.\frac{d\lambda}{d\omega} \right|_{(\lambda_0,\omega_0)}>0,          &  \text{ if }  \omega_0<0.  \\
		\end{aligned}
    \right.
\end{equation}
In other words, when~$\omega_0$ is greater than zero, a small increment of~$\lambda_0$ will decrease the value of~$\omega_0$; and when~$\omega_0$ is less than zero, a small increment of~$\lambda_0$ will increase the value of~$\omega_0$. 
In both cases, when~$\lambda_0$ increases, $|\omega_0|$ will decrease at the new local minimum of~$E(\lambda,\omega)$.

Thus, we finish the proof of Theorem~\ref{theorem:main}. \QEDB

\bibliographystyle{IEEEtran}
\bibliography{references}

\begin{thebibliography}{10}
\providecommand{\url}[1]{#1}
\csname url@samestyle\endcsname
\providecommand{\newblock}{\relax}
\providecommand{\bibinfo}[2]{#2}
\providecommand{\BIBentrySTDinterwordspacing}{\spaceskip=0pt\relax}
\providecommand{\BIBentryALTinterwordstretchfactor}{4}
\providecommand{\BIBentryALTinterwordspacing}{\spaceskip=\fontdimen2\font plus
\BIBentryALTinterwordstretchfactor\fontdimen3\font minus
  \fontdimen4\font\relax}
\providecommand{\BIBforeignlanguage}[2]{{%
\expandafter\ifx\csname l@#1\endcsname\relax
\typeout{** WARNING: IEEEtran.bst: No hyphenation pattern has been}%
\typeout{** loaded for the language `#1'. Using the pattern for}%
\typeout{** the default language instead.}%
\else
\language=\csname l@#1\endcsname
\fi
#2}}
\providecommand{\BIBdecl}{\relax}
\BIBdecl

\bibitem{HanLiuMao16}
S.~Han, X.~Liu, H.~Mao, J.~Pu, A.~Pedram, M.~A. Horowitz, and W.~J. Dally,
  ``{EIE}: Efficient inference engine on compressed deep neural network,'' in
  \emph{ISCA}, 2016.

\bibitem{AnwSun16}
S.~Anwar and W.~Sung, ``Compact deep convolutional neural networks with coarse
  pruning,'' \emph{arXiv preprint arXiv: 1610.09639}, 2016.

\bibitem{SzeCheYanEme17}
V.~Sze, Y.~H. Chen, T.~J. Yang, and J.~Emer, ``Efficient processing of deep
  neural networks: A tutorial and survey,'' \emph{arXiv preprint
  arXiv:1703.09039}, 2017.

\bibitem{lecun1990optimal}
Y.~LeCun, J.~S. Denker, and S.~A. Solla, ``Optimal brain damage,'' in
  \emph{NeurIPS}, 1990.

\bibitem{HasSto93}
B.~Hassibi and D.~G. Stork, ``Second order derivatives for network pruning:
  Optimal brain surgeon,'' in \emph{NeurIPS}, 1993.

\bibitem{MolTyrKar17}
P.~Molchanov, S.~Tyree, and T.~Karras, ``Pruning convolutional neural networks
  for resource efficient inference,'' in \emph{ICLR}, 2017.

\bibitem{VadLem16}
V.~Lebedev and V.~Lempitsky, ``Fast convnets using group-wise brain damage,''
  in \emph{CVPR}, 2016.

\bibitem{wen2016learning}
W.~Wen, C.~Wu, Y.~Wang, Y.~Chen, and H.~Li, ``Learning structured sparsity in
  deep neural networks,'' in \emph{NeurIPS}, 2016.

\bibitem{he2017channel}
Y.~He, X.~Zhang, and J.~Sun, ``Channel pruning for accelerating very deep
  neural networks,'' in \emph{ICCV}, 2017.

\bibitem{yosinski2014transferable}
J.~Yosinski, J.~Clune, Y.~Bengio, and H.~Lipson, ``How transferable are
  features in deep neural networks?'' in \emph{NeurIPS}, 2014.

\bibitem{DinDinHanTan18}
X.~Ding, G.~Ding, J.~Han, and S.~Tang, ``Auto-balanced filter pruning for
  efficient convolutional neural networks,'' in \emph{AAAI}, 2018.

\bibitem{HeZhaRenSun16}
K.~He, X.~Zhang, S.~Ren, and J.~Sun, ``Deep residual learning for image
  recognition,'' in \emph{CVPR}, 2016.

\bibitem{Ree93}
R.~Reed, ``Pruning algorithms -- a survey,'' \emph{{IEEE} Transactions on
  Neural Networks}, vol.~4, no.~5, pp. 740--747, 1993.

\bibitem{HanMaoDal15}
S.~Han, H.~Mao, and W.~Dally, ``{Deep Compression}: Compressing deep neural
  networks with pruning, trained quantization and huffman coding,'' in
  \emph{ICLR}, 2016.

\bibitem{HanTra15}
S.~Han, J.~Pool, J.~Tran, and W.~Dally, ``Learning both weights and connections
  for efficient neural network,'' in \emph{NeurIPS}, 2015.

\bibitem{LiKadDurEtAl17}
H.~Li, A.~Kadav, I.~Durdanovic, H.~Samet, and H.~P. Graf, ``Pruning filters for
  efficient convnets,'' in \emph{ICLR}, 2017.

\bibitem{he2018amc}
Y.~He, J.~Lin, Z.~Liu, H.~Wang, L.-J. Li, and S.~Han, ``{AMC}: {AutoML} for
  model compression and acceleration on mobile devices,'' in \emph{ECCV}, 2018.

\bibitem{Yuan2006Model}
M.~Yuan and Y.~Lin, ``Model selection and estimation in regression with grouped
  variables,'' \emph{Journal of the Royal Statistical Society}, vol.~68, no.~1,
  pp. 49--67, 2006.

\bibitem{HinSriKri12}
G.~Hinton, N.~Srivastava, A.~Krizhevsky, I.~Sutskever, and R.~Salakhutdinov,
  ``Improving neural networks by preventing co-adaptation of feature
  detectors,'' \emph{arXiv preprint arXiv:1207.0580}, 2012.

\bibitem{srivastava2014dropout}
N.~Srivastava, G.~Hinton, A.~Krizhevsky, I.~Sutskever, and R.~Salakhutdinov,
  ``Dropout: A simple way to prevent neural networks from overfitting,''
  \emph{JMLR}, vol.~15, no.~1, pp. 1929--1958, 2014.

\bibitem{wang2017structured}
H.~Wang, Q.~Zhang, Y.~Wang, and H.~Hu, ``Structured probabilistic pruning for
  convolutional neural network acceleration,'' in \emph{BMVC}, 2018.

\bibitem{louizos2017bayesian}
C.~Louizos, K.~Ullrich, and M.~Welling, ``Bayesian compression for deep
  learning,'' in \emph{NeurIPS}, 2017.

\bibitem{molchanov2017variational}
D.~Molchanov, A.~Ashukha, and D.~Vetrov, ``Variational dropout sparsifies deep
  neural networks,'' in \emph{ICML}, 2017.

\bibitem{neklyudov2017structured}
K.~Neklyudov, D.~Molchanov, A.~Ashukha, and D.~Vetrov, ``Structured bayesian
  pruning via log-normal multiplicative noise,'' in \emph{NeurIPS}, 2017.

\bibitem{IanMosAsh16}
F.~Iandola, M.~Moskewicz, and K.~Ashraf, ``{SqueezeNet}: {AlexNet}-level
  accuracy with 50x fewer parameters and $<$0.5{MB} model size,'' \emph{arXiv
  preprint arXiv:1602.07360}, 2016.

\bibitem{Howard2017MobileNets}
A.~G. Howard, M.~Zhu, B.~Chen, D.~Kalenichenko, W.~Wang, T.~Weyand,
  M.~Andreetto, and H.~Adam, ``Mobile{N}ets: Efficient convolutional neural
  networks for mobile vision applications,'' \emph{arXiv preprint
  arXiv:1704.04861}, 2017.

\bibitem{sandler2018mobilenetv2}
M.~Sandler, A.~Howard, M.~Zhu, A.~Zhmoginov, and L.-C. Chen, ``Mobile{N}etv2:
  Inverted residuals and linear bottlenecks,'' in \emph{CVPR}, 2018.

\bibitem{Zhang2017ShuffleNet}
X.~Zhang, X.~Zhou, M.~Lin, and J.~Sun, ``Shuffle{N}et: An extremely efficient
  convolutional neural network for mobile devices,'' \emph{arXiv preprint
  arXiv:1707.01083}, 2017.

\bibitem{ma2018shufflenet}
N.~Ma, X.~Zhang, H.-T. Zheng, and J.~Sun, ``Shuffle{N}et v2: Practical
  guidelines for efficient cnn architecture design,'' in \emph{ECCV}, 2018.

\bibitem{CheWilTyrWeiChe15}
W.~Chen, J.~T. Wilson, S.~Tyree, K.~Q. Weinberger, and Y.~Chen, ``Compressing
  neural networks with the hashing trick,'' in \emph{ICML}, 2015.

\bibitem{CouBen16}
M.~Courbariaux and Y.~Bengio, ``{BinaryNet}: Training deep neural networks with
  weights and activations constrained to~$+1$ or~$-1$,'' \emph{arXiv preprint
  arXiv:1602.02830}, 2016.

\bibitem{LinCouMemBen16}
Z.~Lin, M.~Courbariaux, R.~Memisevic, and Y.~Bengio, ``Neural networks with few
  multiplications,'' \emph{arXiv preprint arXiv:1510.03009}, 2016.

\bibitem{RasOrdRedFar16}
M.~Rastegari, V.~Ordonez, J.~Redmon, and A.~Farhadi, ``{XNOR-net}: Imagenet
  classification using binary convolutional neural networks,'' in \emph{ECCV},
  2016.

\bibitem{DenZarBruLecFer14}
E.~Denton, W.~Zaremba, J.~Bruna, Y.~LeCun, and R.~Fergus, ``Exploiting linear
  structure within convolutional networks for efficient evaluation,'' in
  \emph{NeurIPS}, 2014.

\bibitem{LebYarRakOseLem16}
V.~Lebedev, Y.~Ganin, M.~Rakhuba, I.~Oseledets, and V.~Lempitsky, ``Speeding-up
  convolutional neural networks using fine-tuned {CP}-decomposition,''
  \emph{arXiv preprint arXiv:1510.03009}, 2016.

\bibitem{ZhaZouHeSun16}
X.~Zhang, J.~Zou, K.~He, and J.~Sun, ``Accelerating very deep convolutional
  networks for classification and detection,'' \emph{TPAMI}, vol.~38, no.~10,
  pp. 1943--1955, 2016.

\bibitem{buciluǎ2006model}
C.~Buciluǎ, R.~Caruana, and A.~Niculescu-Mizil, ``Model compression,'' in
  \emph{SIGKDD}, 2006.

\bibitem{ba2014deep}
J.~Ba and R.~Caruana, ``Do deep nets really need to be deep?'' in
  \emph{NeurIPS}, 2014.

\bibitem{hinton2015distilling}
G.~Hinton, O.~Vinyals, and J.~Dean, ``Distilling the knowledge in a neural
  network,'' \emph{arXiv preprint arXiv:1503.02531}, 2015.

\bibitem{zagoruyko2016paying}
S.~Zagoruyko and N.~Komodakis, ``Paying more attention to attention: Improving
  the performance of convolutional neural networks via attention transfer,'' in
  \emph{ICLR}, 2017.

\bibitem{chen2018darkrank}
Y.~Chen, N.~Wang, and Z.~Zhang, ``Darkrank: Accelerating deep metric learning
  via cross sample similarities transfer,'' in \emph{AAAI}, 2018.

\bibitem{ChePurSim06}
K.~Chellapilla, S.~Puri, and P.~Simard, ``High performance convolutional neural
  networks for document processing,'' in \emph{International Workshop on
  Frontiers in Handwriting Recognition}, 2006.

\bibitem{CheWooVan14}
S.~Chetlur, C.~Woolley, and P.~Vandermersch, ``cu{DNN}: Efficient primitives
  for deep learning,'' \emph{arXiv preprint arXiv:1410.0759}, 2014.

\bibitem{hu2016trimming}
H.~Hu, R.~Peng, T.~Yu-Wing, and T.~Chi-Keung, ``Network trimming: A data-driven
  neuron pruning approach towards efficient deep architectures,'' \emph{arXiv
  preprint arXiv:1607.03250}, 2016.

\bibitem{KriHin09}
A.~Krizhevsky, ``Learning multiple layers of features from tiny images,''
  Citeseer, Tech. Rep., 2009.

\bibitem{JiaSheDonEtAl14}
Y.~Jia, E.~Shelhamer, J.~Donahue, S.~Karayev, J.~Long, R.~Girshick,
  S.~Guadarrama, and T.~Darrell, ``Caffe: Convolutional architecture for fast
  feature embedding,'' in \emph{ACM MM}, 2014.

\bibitem{KriSutHin12}
A.~Krizhevsky, I.~Sutskever, and G.~E. Hinton, ``Image{N}et classification with
  deep convolutional neural networks,'' in \emph{NeurIPS}, 2012.

\bibitem{Simonyan2014Very}
K.~Simonyan and A.~Zisserman, ``Very deep convolutional networks for
  large-scale image recognition,'' \emph{Computer Science}, 2014.

\bibitem{liu2017learning}
Z.~Liu, J.~Li, Z.~Shen, G.~Huang, S.~Yan, and C.~Zhang, ``Learning efficient
  convolutional networks through network slimming,'' in \emph{ICCV}, 2017.

\end{thebibliography}
\end{document}